\pdfoutput=1

\documentclass[11pt]{article}

\usepackage[]{acl}

\usepackage{times}
\usepackage{latexsym}

\usepackage[T1]{fontenc}

\usepackage[utf8]{inputenc}

\usepackage{microtype}

\usepackage{amsmath}
\usepackage{amssymb}
\usepackage{amsfonts}
\usepackage{xcolor} 
\usepackage{enumitem}
\usepackage{graphicx}
\graphicspath{{Figure/}}

\title{PLATO-XL: Exploring the Large-scale Pre-training of\\ Dialogue Generation}

\author{Siqi Bao\thanks{~~Equal contribution.}~~~~~ Huang He\footnotemark[1]~~~~~ Fan Wang\footnotemark[1]~~~~~ Hua Wu\footnotemark[1]~~~~~ Haifeng Wang\\
\bf{Wenquan Wu~~~ Zhihua Wu~~~ Zhen Guo~~~ Hua Lu~~~ Xinxian Huang}\\
\bf{Xin Tian~~~~~ Xinchao Xu~~~~~ Yingzhan Lin~~~~~ Zheng-Yu Niu} \\
Baidu Inc., China \\
\texttt{\{baosiqi, hehuang, wang.fan, wu\_hua\}@baidu.com}
}

\begin{document}
\maketitle
\begin{abstract}
To explore the limit of dialogue generation pre-training, we present the models of PLATO-XL with up to 11 billion parameters, trained on both Chinese and English social media conversations. To train such large models, we adopt the architecture of unified transformer with high computation and parameter efficiency. In addition, we carry out multi-party aware pre-training to better distinguish the characteristic information in social media conversations. With such designs, PLATO-XL successfully achieves superior performances as compared to other approaches in both Chinese and English chitchat. We further explore the capacity of PLATO-XL on other conversational tasks, such as knowledge grounded dialogue and task-oriented conversation. The experimental results indicate that PLATO-XL obtains state-of-the-art results across multiple conversational tasks, verifying its potential as a foundation model of conversational AI. 
\end{abstract}

\section{Introduction}
The efficacy of the pre-training paradigm, where large-scale transformer models are trained with massive plain texts, has been widely recognized in natural language processing \citep{devlin2019bert, radford2018improving}. To further boost the performance of these language models, there is a trend to enlarge the model size, dataset size, and the amount of compute used for training \citep{raffel2020exploring, kaplan2020scaling}. Particularly, the GPT-3 model with 175B parameters demonstrates strong zero-shot or few-shot learning capacities without task-specific fine-tuning on downstream tasks \citep{brown2020language}.

Distinct from the general language models, dialogue generation models are usually pre-trained with human-like conversations collected from social media. DialoGPT \citep{zhang2019dialogpt} attempts to train dialogue models with Reddit comments on the basis of pre-trained language models. More recently developed models, like Meena \citep{adiwardana2020towards}, Blender \citep{roller2020recipes}, and PLATO-2 \citep{bao2020plato}, achieve substantial performance improvements on multi-turn conversations. These models have been scaled up to billions of parameters and taken advantage of many more social media conversations for pre-training. Nevertheless, in dialogue generation, there still lacks a clear conclusion about the correlation between model scale and conversation quality. For instance, DialoGPT has three model sizes: 117M, 345M, and 762M, where the 345M one obtains the best performance in their evaluations. Meanwhile, the human evaluations of Blender reveal that the 2.7B model achieves better performance as compared to the one with 9.4B parameters.

In this paper, we argue that the conversation quality may keep benefiting from the enlarged model scale with appropriate pre-training designs. To this end, we explore the large-scale pre-training of dialogue generation models with up to 11B model parameters, namely PLATO-XL. To train such a large model, we adopt the architecture of unified transformer with high computation and parameter efficiency. In addition, we carry out multi-party aware pre-training to better distinguish the characteristic information in social media conversations. With such designs, PLATO-XL achieves superior performances as compared to other approaches in both Chinese and English chitchat. More specifically, PLATO-XL shows a strong capability of absorbing common knowledge within its huge parameters; therefore, it is able to alleviate the well-known hallucination problem\footnotemark[1].
\footnotetext[1]{Generation models might generate some plausible statements with factual errors, also known as "hallucination" problem \citep{marcus2020next}. This problem can be alleviated by expanding model parameters \citep{roberts2020much} or incorporating external non-parametric memories \citep{lewis2020retrieval}.}
Besides, thanks to the multi-party aware pre-training, PLATO-XL effectively reduces the inconsistency phenomenon in multi-turn conversations. 

In addition to open-domain chitchat discussed above, there are two other common conversational tasks \citep{gao2018neural}: knowledge grounded dialogue, and task-oriented conversation. In the experiments, we also explore the ability of PLATO-XL as the foundation model of conversational AI. Our experimental results indicate that PLATO-XL is able to outperform other dialogue generation models across multiple conversational tasks. We have released our source code together with the English model at GitHub\footnotemark[2], hoping to facilitate frontier research in dialogue generation. 
\footnotetext[2]{\url{https://github.com/PaddlePaddle/Knover/tree/develop/projects/PLATO-XL}}

\section{Related Work}
\subsection{Large-scale Pre-trained Language Models}
The pre-training paradigm has brought substantial performance improvements in natural language processing, where large-scale transformer models are pre-trained with massive plain texts. BERT \citep{devlin2019bert} learns to capture the deep bi-directional representation for the input context and achieves remarkable breakthroughs in natural language understanding. GPT \citep{radford2018improving} and GPT-2 \citep{radford2019language} are typical models in natural language generation, which extract uni-directional representation and perform auto-regressive generation. To further boost the performance of language models, there is a trend to enlarge the model size, dataset size, and the amount of compute used for training \citep{raffel2020exploring, kaplan2020scaling}. Particularly, GPT-3 \citep{brown2020language} scales up to 175B parameters and demonstrates strong ability in the zero/few-shot settings. Recently, some larger pre-trained language models are presented with superior performance, including the 178B parameter Jurassic-1 \citep{lieber2021jurassic}, the 280B parameter Gopher \citep{rae2021scaling}, the 530B parameter Megatron-Turing NLG \citep{smith2022using}, and the 540B parameter PaLM \citep{chowdhery2022palm}.

Besides the above English models, there are some large-scale Chinese language models. CPM \citep{zhang2020cpm} maintains a similar model architecture as GPT with 2.6B parameters. CPM-2 \citep{zhang2021cpm} scales up to 11B parameters and employs knowledge inheritance from existing models to accelerate the pre-training process. PanGu-$\alpha$ \citep{zeng2021pangu} is a huge model, with up to 200B parameters. The effective training is carried out on a cluster of 2048 Ascend 910 AI processors with multi-dimension parallelisms and topology-aware scheduling. ERNIE 3.0 \citep{sun2021ernie} proposes a unified framework that integrates both auto-encoding and auto-regressive networks, where knowledge graphs are also encoded into pre-training for enhanced representation. Empirical results show that the 260B parameter ERNIE 3.0 Titan \citep{wang2021ernie} achieves superior performance on 68 Chinese NLP tasks.

\begin{figure*}
	\centering
	\includegraphics[width=0.9\textwidth]{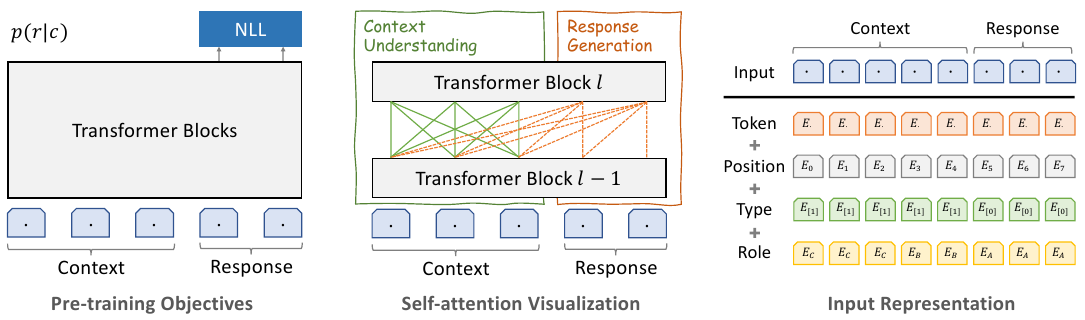}
	\caption{Network overview of PLATO-XL.}
	\label{fig:network}
\end{figure*} 
\subsection{Pre-trained Dialogue Models}
Unlike the plain texts for general language models, for dialogue generation pre-training, human-like conversations are collected from social media, such as Twitter, Reddit, Sina Weibo, Baidu Tieba, etc. DialoGPT \citep{zhang2019dialogpt} attempts to train dialogue models with Reddit comments on the basis of pre-trained language models. Meena \citep{adiwardana2020towards} carries out the pre-training of dialogue generation directly with more social media conversations, and this 2.6B parameter model achieves significant improvements in multi-turn conversation quality. Blender \citep{roller2020recipes} proposes to fine-tune the pre-trained dialogue model with human-annotated datasets to emphasize the conversational skills of engagingness, knowledge, empathy, and personality. In addition, to mitigate the safe response problem, PLATO \citep{bao2019plato} and PLATO-2 \citep{bao2020plato} propose to encode the discrete latent variable into transformer for diverse response generation. Recently, the 137B parameter LaMDA \citep{thoppilan2022lamda} has been introduced particularly for dialogue applications, which is the largest dialogue model in English.

Besides the above English models, PLATO-2 has one Chinese dialogue model of 363 million parameters, exhibiting notable improvements over the classical chatbot of XiaoIce \citep{zhou2020design}. There are some other Chinese dialogue models on a similar modest scale, including CDial-GPT \citep{wang2020large} and ProphetNet-X \citep{qi2021prophetnet}. Recently, one Chinese dialogue model of EVA \citep{zhou2021eva} has been developed under the architecture of Seq2Seq, with up to 2.8B parameters. In this paper, we will introduce the 11B parameter model of PLATO-XL, trained on both Chinese and English social media conversations. To our knowledge, PLATO-XL is the largest pre-trained dialogue model in Chinese so far.

\section{PLATO-XL}
\subsection{Network Overview}
The network overview of PLATO-XL is shown in Figure \ref{fig:network}, with transformer blocks as the backbone. For the sake of efficient training on a large scale, PLATO-XL keeps the adoption of the unified transformer \citep{bao2019plato, bao2020plato} (also known as PrefixLM \citep{raffel2020exploring, dong2019unified}) instead of the typical encoder-decoder for dialogue generation. The advantages brought by the unified transformer architecture are two-fold: computation and parameter efficiency. Firstly, given the conversation samples of variable lengths, it is necessary to pad them into a certain length in the training process, which inevitably incurs massive invalid computations. As suggested in fairseq \citep{ott2019fairseq}, the amount of padding can be minimized by grouping the input with similar lengths. By performing effective sorting on the concatenated input, invalid computations caused by padding can be reduced significantly with the unified transformer. Secondly, through the flexible mechanism of the self-attention mask, the two tasks of dialogue context understanding and response generation are modeled simultaneously with shared parameters. As such, the unified transformer is more parameter-efficient than the encoder-decoder network \citep{bao2020plato, du2021all}. 

In PLATO-XL, the pre-training objective is to minimize the negative log-likelihood (NLL) loss:
\begin{equation}\nonumber
\begin{split}
\mathcal{L}_{NLL}&=-\mathbb{E}_{(c,r)\sim D}~\left[\log~ p_\theta (r|c)\right]\\
&=-\mathbb{E}_{(c,r)\sim D}~\left[\sum_{t=1}^T~\log~ p_\theta(r_t|c,r_{<t})\right]~,
\end{split}
\end{equation}
where $\theta$ refers to the trainable parameters of the dialogue generation model and $D$ stands for the pre-training data. The input to the network is a pair of dialogue context $c$ and target response $r$. $T$ is the length of the target response and $r_{<t}$ denotes previously generated words. As shown in Figure \ref{fig:network}, the input representation is calculated as the sum of the corresponding token, position, type, and role embeddings. The token and position embeddings are commonly used in pre-training models. The type embedding is employed to differentiate the segments of dialogue context and target response, which is also extensible for other input sources, such as persona profiles or grounded knowledge used in conversations. The role embedding is used to distinguish the characters in the multi-turn conversations, which will be explained in detail in the following subsection.

\begin{figure*}
	\centering
	\includegraphics[width=0.9\textwidth]{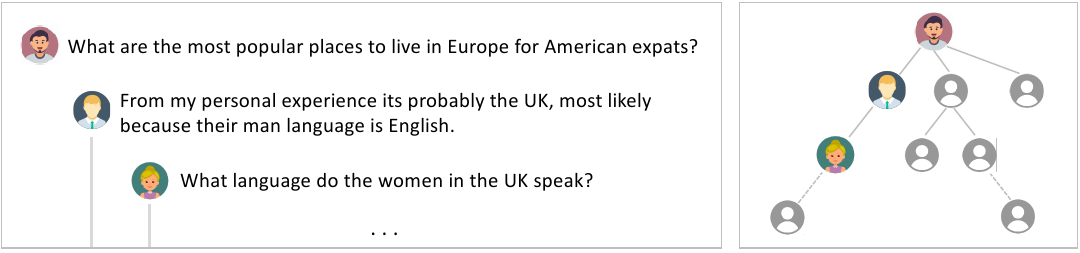}
	\caption{Left: one toy example to illustrate social media conversations. Right: corresponding message tree.}
	\label{fig:social}
\end{figure*} 
\subsection{Multi-Party Aware Pre-training}
As discussed in the related work, general language models are pre-trained with massive plain texts, where each training sample is usually created by one single author or user. In comparison, the dialogue models are commonly pre-trained with human-like conversations collected from public social media, where one toy example is provided in Figure \ref{fig:social} for illustration. Several properties of social media conversations can be observed from this example: 1) there are multi-level comments appended to respond to the contexts; 2) multiple users are actively involved in the discussion. The corresponding message tree of these comments is shown on the right-hand side. The comments along the path from the root node to any tree node can be formulated as one training sample of dialogue context and target response. However, with these social media conversations, the learned models tend to mix information from multiple characters in the context and have difficulties generating consistent responses.

To tackle the above problem, PLATO \citep{bao2019plato} first introduces the role embedding into the transformer to distinguish the characters in the dialogue context. While there is an underlying assumption in PLATO that the conversation is carried out within two characters and the role embedding is assigned alternatively. Although it is generally tenable in human-annotated conversations, things get complicated with social media conversations. As suggested in the former works of RNN-based response selection \citep{ouchi2016addressee, zhang2018addressee}, user embedding is an effective technique for speaker and addressee identification in multi-party conversation. In PLATO-XL, we further encode the multi-party aware role embedding in the pre-training of dialogue generation. The target response and utterances in the context by the same user will be assigned with the role embedding of $E_A$. For the rest utterances, the role embedding will be assigned in a relative order according to the user ids, such as $E_B$, $E_C$, etc. This multi-party aware pre-training helps the model distinguish the information in the context and maintain consistency in dialogue generation.

\subsection{Pre-training Settings}
For the pre-training corpora, the English conversation samples are extracted from Reddit comments, which are collected by a third party and made publicly available at pushshift.io \citep{baumgartner2020pushshift}. To guarantee the data quality, we follow the elaborate cleaning process as PLATO-2 \citep{bao2020plato}. After filtering, the data is split into training and validation sets in chronological order. The training set contains 811M (context, response) samples, ranging from December 2005 to December 2019. For the validation set, 0.2M samples are selected from the rest data after December 2019. The English vocabulary contains 8K BPE tokens \citep{sennrich2016neural}, constructed with the SentencePiece library. The Chinese pre-training data is collected from public domain social media. After filtering, there are 1.2B (context, response) samples in the training set. As for the Chinese vocabulary, it contains 30K BPE tokens.

PLATO-XL employs the same network architecture for the Chinese and English models, with up to 11 billion parameters. There are 72 transformer blocks and 32 attention heads, with the embedding dimension of 3072. The hidden dimension of the feedforward layer is set to 18432. Pre-normalization connection and scaled initialization \citep{radford2019language} are adopted for stable training. The main hyper-parameters used in the pre-training are listed as follows. The maximum sequence length for the dialogue context and target response is set to 896 and 128, respectively. We use Adam \citep{kingma2015adam} as the optimizer with a learning rate scheduler of linear warmup and decay. The warmup stage covers the first 200 steps, and the peak learning rate is 8e-5.   

The implementation of PLATO-XL is based on the PaddlePaddle platform. And the training was carried out on 256 Nvidia Tesla V100 32G GPU cards. Given the limited memory of each device, vanilla data parallelism cannot support the training of such a model with up to 11 billion parameters. As such, we adopt the sharded data parallelism \citep{rajbhandari2020zero} to eliminate memory redundancies by partitioning the optimizer states, gradients, and parameters across multiple devices. This kind of distributed training helps maintain low communication volume and high computational granularity. In addition, to train the model with a relatively large batch size, we further employ gradient checkpointing \citep{chen2016training} to trade computation for memory. In PLATO-XL, each model was trained for a total of 150B tokens, with a batch size of 2M tokens.

\section{Experiments}
\subsection{Evaluation Settings}

\subsubsection{Compared Approaches}
To evaluate the performance of PLATO-XL, we compare it with the following English and Chinese dialogue generation models in the experiments.
\begin{itemize}[leftmargin=*,noitemsep,topsep=0pt]
    \item DialoGPT \citep{zhang2019dialogpt} is trained on the basis of GPT-2 \citep{radford2019language} using Reddit comments. There are three model sizes: 117M, 345M, and 762M. Since the 345M parameter model obtains the best performance in their evaluations, this version is compared.
    
    \item Blender \citep{roller2020recipes} is first trained using Reddit comments and then fine-tuned with human-annotated conversations -- BST \citep{smith2020can}, to help emphasize desirable conversational skills of engagingness, knowledge, empathy, and personality. Blender has three model sizes: 90M, 2.7B, and 9.4B. Since the 2.7B parameter model obtains the best performance in their evaluations, this version is compared.
    
    \item PLATO-2 \citep{bao2020plato} is trained via curriculum learning, where a coarse-grained model is first learned for general response generation and a fine-grained model is further learned for diverse response generation. The English model of PLATO-2 is pre-trained with Reddit comments and then fine-tuned with BST conversations. There are 1.6B parameters in this model. PLATO-2 also has one Chinese model of 336M parameters, trained with 1.2B social media conversation samples. 

    \item CDial-GPT \citep{wang2020large} is trained on the basis of a Chinese GPT model using LCCC conversations. There are 95.5M parameters in this model.
    
    \item ProphetNet-X \citep{qi2021prophetnet} is a family of pre-trained models on various languages and domains. ProphetNet-X includes one Chinese dialogue generation model trained on social media conversations collected from Douban group\footnotemark[3]. There are 379M parameters in this model.
    
    \item EVA \citep{zhou2021eva} is a 2.8B parameter Chinese dialogue generation model trained with the WDC-Dialogue, which includes 1.4B conversation samples collected from social media. 
\end{itemize}
\footnotetext[3]{\url{https://www.douban.com/group/}}

In addition to the above models, PLATO-XL is also compared with the following commercial chatbots in Chinese:
Microsoft XiaoIce \citep{zhou2020design}, Turing Robot\footnotemark[4], Tmall Genie\footnotemark[5], and Xiao AI\footnotemark[6].\footnotetext[4]{\url{http://www.turingapi.com/}}\footnotetext[5]{\url{https://bot.tmall.com/}}\footnotetext[6]{\url{https://xiaoai.mi.com/}} The official platform/API is used in the interactions with XiaoIce and Turing. As there is no public API for Tmall Genie or Xiao AI, voice interactions are carried out instead with these smart speakers. 

\subsubsection{Evaluation Metrics}
As suggested in the empirical study \citep{liu2016not}, the correlation between automatic metrics and human judgments is weak in open-domain dialogue generation. Therefore, we mainly rely on human evaluations in the experiments of open-domain conversation. Crowd-sourcing workers are asked to evaluate the conversation quality on the following aspects. 
\begin{itemize}[leftmargin=*,noitemsep,topsep=0pt]
	\item Coherence is an utterance-level metric, measuring whether the response is relevant and consistent with the context. 
	
	\item Informativeness is also an utterance-level metric, evaluating whether the response is informative or not given the context.
	
	\item Engagingness is a dialogue-level metric, assessing whether the annotator would like to talk with the speaker for a long conversation.
\end{itemize}
The scale of the above metrics is [0, 1, 2]. The higher score, the better. To further analyze the conversation quality, two more fine-grained metrics are included in the evaluation. 
\begin{itemize}[leftmargin=*,noitemsep,topsep=0pt]
	\item Inconsistency is one fine-grained metric for coherence evaluation, checking whether the response conflicts with the context. 
	
	\item Hallucination is one fine-grained metric for informativeness evaluation, checking whether the response contains any factual errors.
\end{itemize}
The scale of inconsistency and hallucination is [0, 1]. The lower score, the better. Score details about these metrics are provided in the Appendix.

\begin{table*}
	\centering
	\includegraphics[width=0.9\textwidth]{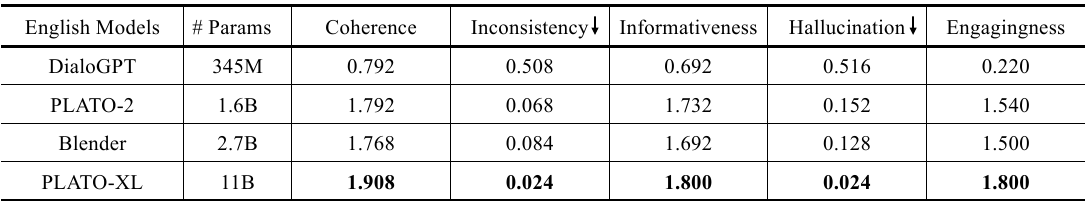}
	\caption{English self-chat evaluation results, with best value written in bold.}
	\label{tab:self-chat-en}
\end{table*} 
\begin{table*}
	\centering
	\includegraphics[width=0.9\textwidth]{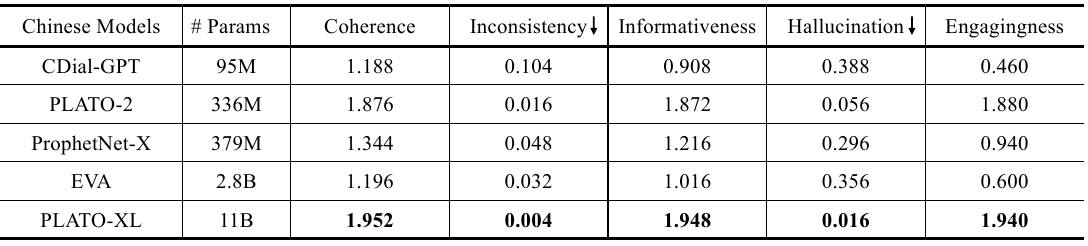}
	\caption{Chinese self-chat evaluation results, with best value written in bold.}
	\label{tab:self-chat-zh}
\end{table*} 
\begin{table*}
	\centering
	\includegraphics[width=0.9\textwidth]{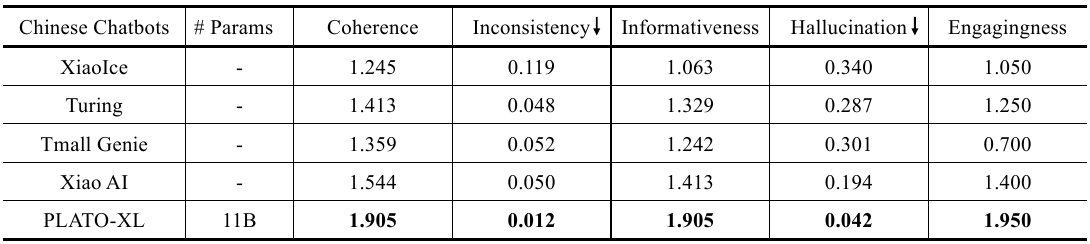}
	\caption{Chinese human-bot chat evaluation results, with best value written in bold.}
	\label{tab:interactive-zh}
\end{table*} 
\subsection{Experimental Results}
\subsubsection{Self-Chat Evaluation}
Self-chats have been widely used in the evaluation of dialogue systems \citep{li2016deep, bao2019know, roller2020recipes}, where a model plays the role of both partners in the conversation. Following the experimental settings in PLATO-2, the interactive conversation is started with a randomly selected topic, and the model performs self-chats for five rounds. Then 50 conversations are selected and distributed to crowd-sourcing workers for evaluation. Each conversation is evaluated by three annotators, and the final score is determined through majority voting. The English and Chinese self-chat evaluation results are summarized in Table \ref{tab:self-chat-en} and \ref{tab:self-chat-zh}, respectively. These results indicate that PLATO-XL is able to produce coherent, informative, and engaging conversations. Particularly, both the inconsistency and hallucination problems of dialogue generation are alleviated remarkably with PLATO-XL. As compared to other approaches, the 11B parameter model achieves superior performances in both Chinese and English chitchat. 

\begin{figure*}
	\centering
	\includegraphics[width=0.98\textwidth]{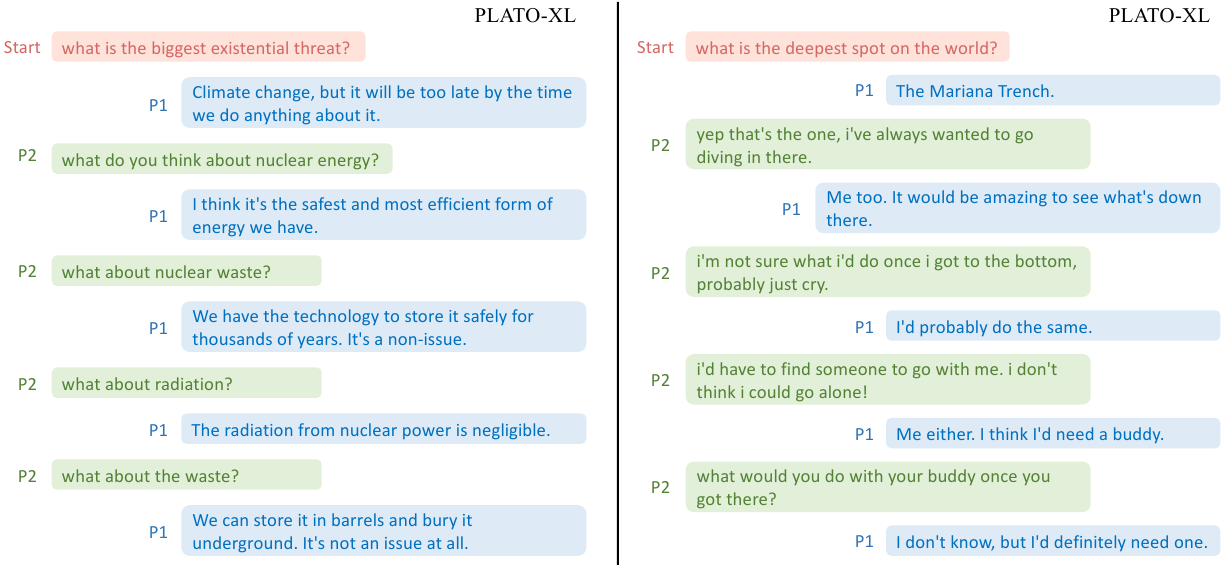}
	\caption{Cherry-picked English self-chat examples by PLATO-XL.}
	\label{fig:self-chat-en-case}
\end{figure*} 
\begin{figure*}
	\centering
	\includegraphics[width=0.98\textwidth]{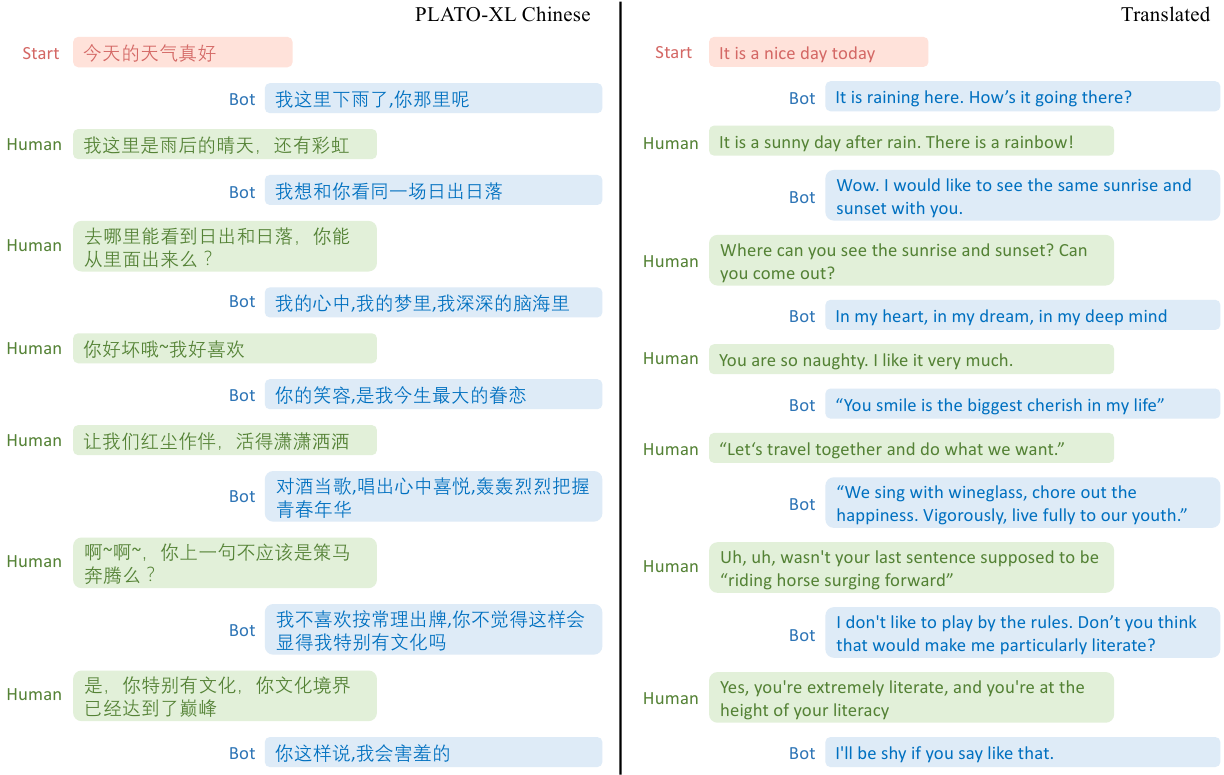}
	\caption{Cherry-picked Chinese human-bot chat example by PLATO-XL.}
	\label{fig:interactive-zh-case}
\end{figure*} 
\subsubsection{Human-Bot Chat Evaluation}
Besides the above public models, PLATO-XL is compared with the following commercial chatbots in Chinese: Microsoft XiaoIce, Turing Robot, Tmall Genie, and Xiao AI. As most of them do not have publicly available APIs, we ask our in-house annotation team to collect the human-bot conversations. The interactive conversation also starts with a pre-selected topic and continues for 7-14 rounds. 20 diverse topics are extracted from the high-frequency topics of a commercial chatbot, including travel, movie, hobby, and so on. The collected human-bot conversations are distributed to crowd-sourcing workers for evaluation. The human-bot chat evaluation results are summarized in Table \ref{tab:interactive-zh}. These results indicate that PLATO-XL achieves significant improvements over the rest of the commercial chatbots across all the human evaluation metrics. 

\subsubsection{Case Analysis}
To further analyze the model's features, two English self-chat examples by PLATO-XL are provided in Figure \ref{fig:self-chat-en-case}. These examples demonstrate that PLATO-XL is able to conduct coherent, informative, and engaging conversations. The in-depth discussions on nuclear energy and Mariana Trench indicate that massive knowledge has been absorbed implicitly in the tremendous parameters. Moreover, from the self-chat example on the left-hand side, it can be observed that the model maintains well the characteristics of each participant. P2 seems like a curious learner, tending to ask many questions. P1 is a knowledgeable expert, providing the answers in detail but with a little impatience. The model is capable of generating responses with good consistency on content and style, thanks to the multi-party aware pre-training. 

One Chinese human-bot chat example by PLATO-XL is provided in Figure \ref{fig:interactive-zh-case}, with original interactive logs shown on the left and translated logs on the right. In this example, PLATO-XL even exhibits advanced conversational skills, such as compliment and eloquence. The model replies to the other partner with sweet words from romantic lyrics and provides reasonable explanations to the queries.  

\begin{table*}
	\centering
	\includegraphics[width=0.9\textwidth]{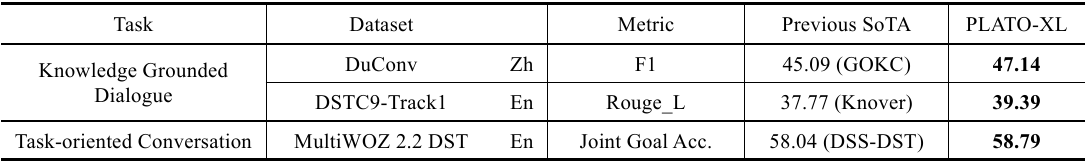}
	\caption{Automatic evaluation results on knowledge grounded and task-oriented conversations, with best value written in bold.}
	\label{tab:other-tasks}
\end{table*} 
\subsection{Explorations on other Conversational Tasks}
In addition to open-domain chitchat, there are two other common conversational tasks \citep{gao2018neural}: knowledge grounded dialogue, and task-oriented conversation. As such, in the experiments, we also explore the ability of PLATO-XL on these conversational tasks.

\subsubsection{Task Descriptions}
The experiments are carried out on the following conversational tasks:
\begin{itemize}[leftmargin=*,noitemsep,topsep=0pt]
    \item DuConv \citep{wu2019proactive} is one Chinese knowledge grounded conversation dataset collected in LUGE\footnotemark[7]. DuConv focuses on proactive conversations towards pre-defined goals and includes 30K dialogues based on movie knowledge graphs.
    
    \item DSTC9-Track1 \citep{kim2020beyond} aims to incorporate external knowledge resources to reply user's out-of-API-coverage queries and augments the dataset of MultiWOZ 2.1 \citep{eric2020multiwoz} with 22K knowledge grounded conversation turns. There are three tasks in DSTC9-Track1: knowledge-seeking turn detection, knowledge selection, and knowledge-grounded response generation. In the experiments, we consider the task of knowledge-grounded response generation. 
    
    \item MultiWOZ 2.2 \citep{zang2020multiwoz} is a polished version of MultiWOZ 2.1, including 10K task-oriented conversations across multiple domains. In the experiments, we consider the classical task of dialog state tracking (DST).
\end{itemize}
\footnotetext[7]{LUGE, Language Understanding and Generation Evaluation Benchmarks, \url{https://www.luge.ai/}}

\subsubsection{Automatic Evaluation}
The fine-tuning experiments of PLATO-XL are carried out on these conversational tasks, with automatic evaluation results summarized in Table \ref{tab:other-tasks}.
\begin{itemize}[leftmargin=*,noitemsep,topsep=0pt]
    \item In DuConv, the model needs to generate the response given related knowledge triplets and lead the conversation to a pre-defined goal. By expanding the network input of PLATO-XL, the conversational goal and knowledge triplets can be easily encoded and grounded for response generation. Compared to the previous state-of-the-art approach -- GOKC \citep{bai2021learning}, PLATO-XL improves the F1 value by 2.05 points.
    
    \item In DSTC9-Track1, we focus on the evaluation of knowledge grounded response generation. In the experiments, we train and test the models with golden retrieved knowledge snippets. The winner approach in DSTC9-Track1 -- Knover \citep{he2021learning}, is also developed on pre-trained dialogue models. The comparison reveals that PLATO-XL further improves the performance by 1.62 points.
    
    \item In MultiWOZ 2.2, PLATO-XL learns to generate the dialog state directly given the context. Compared to the previous SoTA approach -- DSS-DST \citep{guo2021dual}, PLATO-XL further improves the joint goal accuracy to 58.79.  
\end{itemize}
The superior performance of PLATO-XL on multiple conversational tasks verifies its potential as a foundation model of conversational AI.

\section{Conclusion}
In this paper, we explore the large-scale pre-training of dialogue generation and present the 11 billion parameter model of PLATO-XL. Experimental results demonstrate that PLATO-XL achieves superior performance as compared with other approaches in both Chinese and English chitchat. Particularly, the problems of hallucination and inconsistency are alleviated remarkably in PLATO-XL, mainly attributed to the implicit knowledge absorbed in the tremendous parameters and the multi-party aware pre-training. Besides the open-domain conversation, PLATO-XL obtains state-of-the-art results on multiple knowledge grounded and task-oriented conversations, verifying its capacity as a foundation model of conversational AI. 

\section{Ethical Considerations}
With the development of large-scale pre-training models, there raise several ethical concerns, including toxic and biased language. In PLATO-XL, several strategies are explored to boost the safety of open-domain chatbots. In the pre-processing stage, elaborate data cleaning is carried out to remove offensive messages from the training corpora. In the post-processing stage, we employ one classifier to detect sensitive topics from users' utterances and will return canned responses for these contexts. We adopt another classifier to filter out potentially unsafe candidates from generated responses. Moreover, we carry out regular adversarial tests with our in-house data specialists and update the safety classifiers with newly collected samples. Given that the objectives of safety differ across language contexts, we design and employ corresponding strategies for English and Chinese conversations. While even with these strategies, the bot might still generate biased or unsafe statements under sensitive topics or adversarial contexts. Future work will put more emphasis on the fairness and safety of open-domain chatbots. 

\section*{Acknowledgments}
We would like to thank Jingzhou He, Tingting Li, and Shiwei Huang for the help on resource coordination; Jianzhong Liang, and Long Li for the support on PaddlePaddle implementation; Baotong Luo, and Dou Hong for the assistance with infrastructure. This work was supported by the Natural Key Research and Development Project of China (No. 2018AAA0101900).

\bibliography{bibtex}

\clearpage
\appendix
\section{Scoring Criteria in Human Evaluation}
The criteria used in human evaluation are provided in Table \ref{tab:criteria}.
\begin{table}[ht]
	\centering
	\includegraphics[width=\columnwidth]{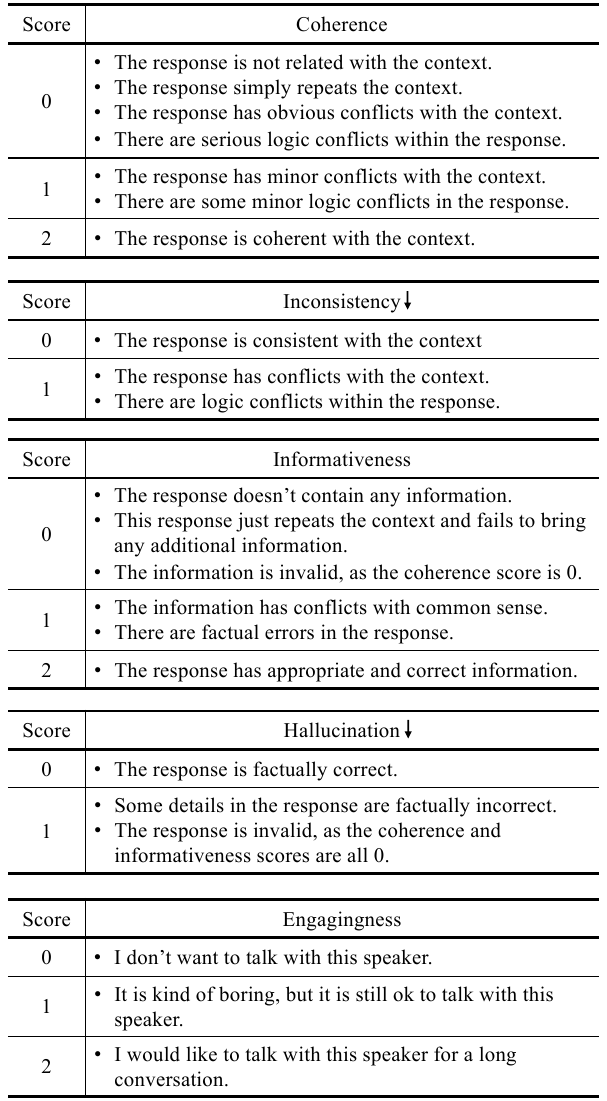}
	\caption{Score details of metrics used in human evaluation.}
	\label{tab:criteria}
\end{table}

\section{Prompting Efficient Dialogue Generation}
In the practical deployment of the large-scale pre-trained dialogue model, one hindrance is the limited inference efficiency. Firstly, the model has tremendous parameters, leading to expensive computational costs. Secondly, in response generation, the model has to generate the response sequence step by step, suffering from high latency. We have explored several strategies to boost inference efficiency, including operation fusion, FP16 computation, and so on. With these techniques, on the Nvidia Tesla V100 32G GPU card, the average latency of 11B parameter Chinese PLATO-XL is successfully reduced to 941ms from 3.3s, resulting in 3.5 times acceleration. To facilitate the deployment of dialogue models, we also have plans to release these acceleration implementations soon.

\end{document}